\documentclass[letterpaper,10pt,conference]{ieeeconf}
\IEEEoverridecommandlockouts
\usepackage{cite}
\usepackage{graphicx}
\usepackage[font=footnotesize]{subfig}
\usepackage{amsmath,amssymb,bm}
\usepackage[table,xcdraw]{xcolor}
\usepackage{multirow}
\usepackage{caption}
\usepackage{booktabs}
\usepackage[colorlinks=true,allcolors=blue]{hyperref}

\begin{document}

\title{\LARGE \bf
A Digital Twin Framework for Virtual Visuo-Haptic Teleoperation of Complex-Shaped Optical Microrobots}

\author{
Zongcai Tan, Lan Wei, Dandan Zhang
\thanks{Zongcai Tan, Lan Wei, and Dandan Zhang are with the Department of Bioengineering, Imperial-X AI Initiative, Imperial College London, London, United Kingdom.  Corresponding: d.zhang17@imperial.ac.uk.  Code is available \href{https://github.com/Zongcai23/ot-haptic-feedback-digital-twin/tree/main}{here}.}
}

\maketitle

\begin{abstract}
Optical tweezers (OT) provide piconewton-scale manipulation for delicate biomedical tasks, where visuo-haptic feedback can improve operator awareness by conveying interaction-force cues and trap-stability information. However, visuo-haptic teleoperation frameworks for complex-shaped optical microrobots remain underdeveloped, particularly in multi-trap manipulation scenarios.
This paper presents a digital twin framework for virtual visuo-haptic teleoperation of complex-shaped OT-driven microrobots. The framework integrates a digital twin environment, image-based pose and depth estimation, microrobot motion simulation, and model-based haptic rendering within a Robot Operating System (ROS)-connected bimanual teleoperation system. 
For force modeling, we combine a Multi-Sphere Distributed Manipulation (MSDM) model with optical-force estimation from the Optical Tweezers Toolbox, enabling simulator-driven visuo-haptic feedback. The framework reproduces representative microrobot motion trends and provides haptic force rendering that is numerically consistent with the fitted optical-force model. In simulated cell-delivery tasks, haptic feedback reduced the standard deviations of the contact-force metric and the microrobot-to-trap-center distance metric by 53.2\% and 55.2\%, respectively, and improved task success from 30\% to 80\%. These results demonstrate the framework’s effectiveness for evaluating visuo-haptic teleoperation strategies for complex-shaped optical microrobots.
\end{abstract}

\begin{figure*}[htbp!]
  \captionsetup{font=footnotesize,labelsep=period}
\centering
\includegraphics[width=0.9\hsize]{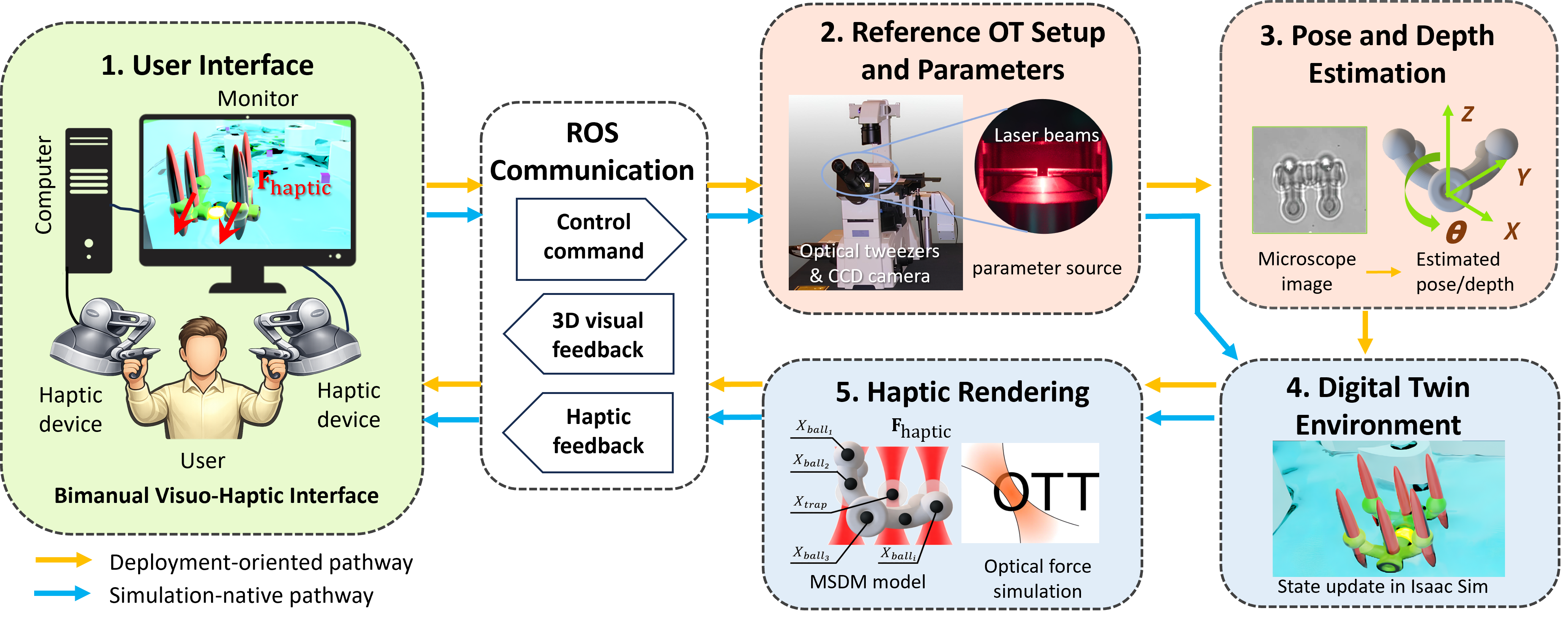}
  \vspace{-0.2cm}
\caption[Workflow of the proposed visuo-haptic teleoperation framework]{Overview of the proposed digital twin framework with two operating pathways. In the deployment-oriented pathway, OT images and parameter settings from the reference setup support pose/depth estimation and state alignment for digital-twin reconstruction toward future physical deployment. In the simulation-native pathway used in this study, user commands directly update the digital twin for motion simulation, visualization, and haptic rendering.}
\label{fig-workflow}
    \vspace{-0.6cm}
\end{figure*}%

\section{Introduction}

Optical tweezers (OT) have been widely used in biomedical applications such as cell manipulation, tissue engineering, and micro-assembly \cite{hu2022advanced,zhang2022fabrication,yadav2024optical}. Despite their capability for non-contact and piconewton-scale manipulation, direct optical interaction with sensitive biological samples can introduce photothermal and phototoxic risks, while direct trapping of target cells may also limit manipulation flexibility \cite{volpe2023roadmap}. To mitigate these limitations, microrobots have been introduced as OT-driven end-effectors for indirect cell manipulation \cite{zhang2022fabrication}. By mediating the interaction between optical traps and biological targets, microrobots can reduce direct laser exposure to cells and provide greater flexibility for manipulating delicate or irregular samples \cite{pan2026optical}.

Nevertheless, indirect manipulation with OT-driven microrobots introduces additional control challenges. Excessive contact forces, high trapping velocities, trap instability, and unintended collisions may cause cell loss during transportation or lead to manipulation failure. These risks are further amplified by the limited depth perception of conventional 2D microscopic visual feedback, which makes it difficult for operators to accurately assess the spatial relationships among optical traps, microrobots, and biological targets. Integrating haptic feedback with real-time 3D pose visualization can therefore enhance operator awareness of both interaction forces and spatial relationships, supporting safer and more precise micromanipulation \cite{fan2023digital}.

Haptic feedback has therefore become increasingly important in biomedical micromanipulation tasks involving cells, tissues, and other delicate biological samples \cite{guo2024lightweight,riaziat2024investigating,si2024enabling}. Recent OT-based haptic systems have demonstrated the potential of force feedback for improving operator perception and manipulation control \cite{gerena20233d,tanaka2022dual}. However, many existing approaches rely on simplified or locally calibrated optical-force models, whose accuracy may decrease as the manipulated object moves away from the trap center or enters more complex trap-object configurations. In addition, most visuo-haptic OT studies have focused on relatively simple objects or manipulation scenarios \cite{raphalen2025haptic}, whereas visuo-haptic teleoperation strategies for complex-shaped optical microrobots remain underexplored.

For complex-shaped optical microrobots, effective teleoperation requires both accurate spatial reconstruction and reliable force estimation. This requirement becomes particularly important in multi-trap manipulation, where operators must understand the spatial relationships among the microrobot, optical traps, and biological targets while receiving meaningful force cues for feedback and control. A digital twin is well suited to this setting because it provides a unified environment for 3D visualization, optical-force modeling, interaction analysis, and operator-in-the-loop evaluation \cite{grieves2016digital}. It can also serve as a low-cost and repeatable testbed for developing visuo-haptic teleoperation strategies before deployment on physical OT platforms.

Recent studies have introduced simulation environments for OT-driven microrobots \cite{tan2025interactive} and perception-oriented digital reconstruction methods based on sim-to-real pose estimation and physics-informed microscope-image generation \cite{tan2025physics}. Related digital-twin and haptic-rendering frameworks have also been explored for magnetic microrobots \cite{keshmiri2022digital,lee2021real}. However, integrated visuo-haptic digital-twin frameworks for OT-actuated complex-shaped microrobots remain largely unexplored, especially for multi-trap manipulation requiring coupled force modeling, spatial reconstruction, and haptic rendering.

To address this gap, we present a digital twin framework for virtual visuo-haptic teleoperation of OT-actuated complex-shaped microrobots. The framework integrates optical-force modeling, microrobot motion simulation, 3D visual reconstruction, model-based haptic rendering, and pose-and-depth estimation to support both simulation-native evaluation and deployment-oriented alignment with OT observations. It provides a cost-effective platform for developing and user-testing teleoperation strategies with enhanced force and spatial awareness. The \textbf{main contributions} are as follows:

\begin{enumerate}
\item An integrated visuo-haptic digital-twin framework for complex-shaped optical microrobots, combining bimanual teleoperation, microrobot motion simulation, 3D visualization, and model-based haptic rendering within a unified ROS-connected environment.
\item A learning-based pose and depth estimation module for aligning microscope-based OT observations with digital-twin reconstruction, enabling estimation of the spatial relationship between the microrobot and optical traps.
\end{enumerate}


\label{methodology}
\begin{figure*}[t!]
  \captionsetup{font=footnotesize,labelsep=period}
\centering
\includegraphics[width=0.85\hsize]{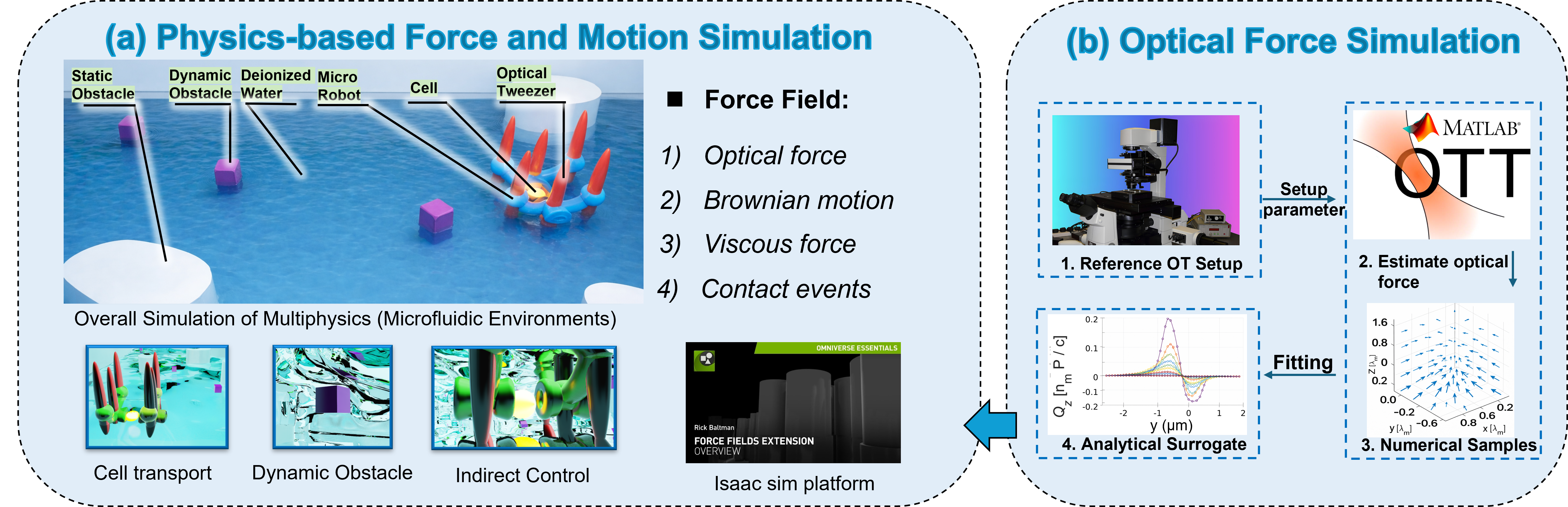}
\caption[Simulated force field environment and modeling method]{Digital twin scene and optical-force surrogate. (a) Isaac Sim scene for microfluidic manipulation, including simulated cells and obstacles. (b) Optical-force samples generated using the Optical Tweezers Toolbox (OTT) \cite{nieminen2007optical} are approximated by a piecewise surrogate model to support efficient motion simulation and haptic rendering.}
\label{fig-forceConsctruct}
    \vspace{-0.2cm}
\end{figure*}%
 
\section{Methodology}
\subsection{System Overview and Workflow}

As shown in Fig.~\ref{fig-workflow}, the framework comprises three components: (1) a user interface with two haptic devices (Geomagic Touch, 3D Systems, USA), (2) a reference OT setup that provides microscopic images and OT-related parameter settings, and (3) a digital twin implemented in Isaac Sim (NVIDIA Omniverse, USA). Communication among these components is handled through the Robot Operating System (ROS).

The digital twin serves as the core environment for microrobot motion simulation, state updating, visual reconstruction, and haptic rendering. In the simulation-native pathway used in this study, bimanual user commands are transmitted through ROS to update the trap configuration and microrobot state directly in the digital twin. The updated state then drives scene reconstruction, visualization, and haptic rendering. In the deployment-oriented pathway, microscopic images from the reference OT setup can be processed by the perception module to estimate the microrobot pose and depth, which are then mapped to the digital twin for state alignment and reconstruction. In both pathways, the reconstructed 3D scene is displayed on the monitor, while model-based haptic feedback is delivered through the two haptic devices.



\subsection{Complex Physical Force Field Simulation in Digital Twin}

As illustrated in Fig.~\ref{fig-forceConsctruct}(a), the digital twin contains a task-level microfluidic scene for OT-actuated microrobot manipulation. The simulation includes the main interaction effects considered in this study, namely optical force, Brownian perturbation, viscous drag, and contact interactions with cells and obstacles. Buoyancy and gravity are neglected because their net contribution is small under the present setting.


For optical-force modeling, numerical force samples are generated using the Optical Tweezers Toolbox (OTT) \cite{nieminen2007optical}, with medium properties, optical beam parameters, trap configurations, and the material and geometric parameters of the MSDM spherical trapping elements derived from the reference OT setup.
As shown in Fig.~\ref{fig-forceConsctruct}(b), these samples are approximated by a piecewise closed-form force model for efficient use in the digital twin. The resulting model is expressed in Eq.~\eqref{eq:optical_force}, which enables real-time optical-force estimation without repeatedly querying the full OTT solution during interactive simulation and haptic rendering.




\begin{figure*}[htbp!]
  \captionsetup{font=footnotesize,labelsep=period}
\centering
\includegraphics[width=0.7\hsize]{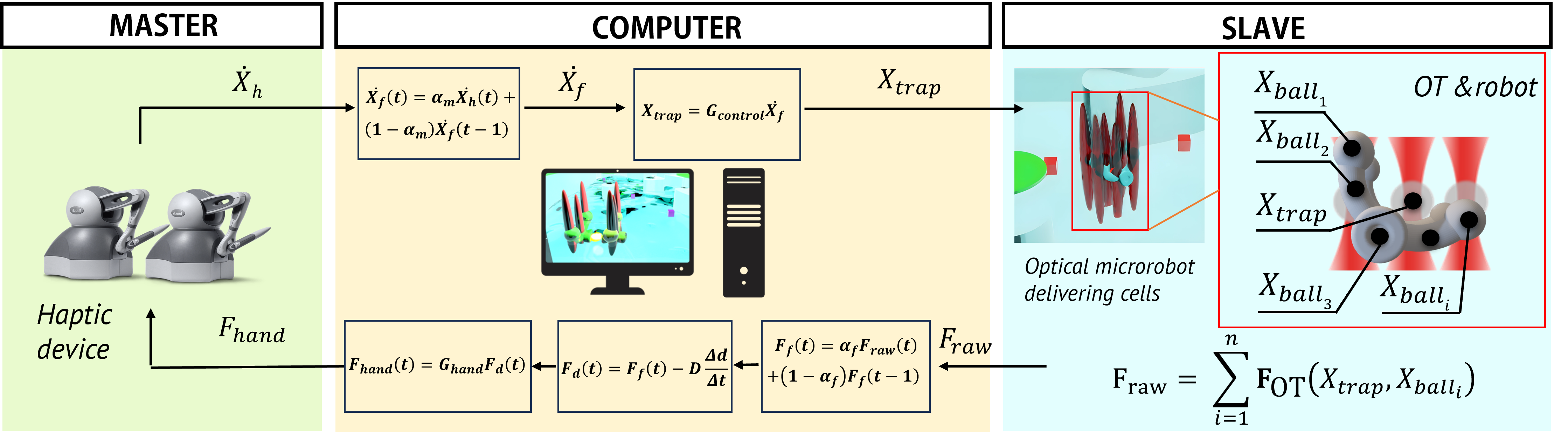}
\caption{Bilateral signal flow of the proposed visuo-haptic OT framework. 
In the forward path, the hand-motion input $\dot{X}_{h}$ is smoothed by a discrete low-pass filter ($\alpha_m=0.05$) and mapped by $G_{\mathrm{control}}$ to the optical-trap position $X_{\mathrm{trap}}$. 
In the feedback path, the trap--microrobot geometry and fitted OT force model estimate the raw model-based force $F_{\mathrm{raw}}$, which is then filtered ($\alpha_f=0.05$), virtually damped, and scaled by $G_{\mathrm{hand}}$ ($G_{\mathrm{hand}}=0.0022$--$0.0030$) to generate the rendered hand force $F_{\mathrm{hand}}$. 
Here, the subscripts $h$, $f$, and $d$ denote the hand input, filtered signal, and virtually damped signal, respectively. 
The inset shows the MSDM-based spherical approximation and trap-relative geometry used for optical-force computation.}

\label{fig-controlBlock}
    \vspace{-0.4cm}
\end{figure*}%

\begin{equation}
\resizebox{\linewidth}{!}{$
\mathbf{F}_{\text{OT}}(X_{\text{trap}}, X_{\text{object}}) = 
\begin{cases}
K \cdot \| X_{\text{trap}} - X_{\text{object}} \|, & \text{if} \ \| X_{\text{trap}} - X_{\text{object}} \| < \delta \\
C + \frac{A}{\| X_{\text{trap}} - X_{\text{object}} \|^2}, & \text{if} \ \| X_{\text{trap}} - X_{\text{object}} \| \geq \delta
\end{cases}
$}
\label{eq:optical_force}
\end{equation}

Here, $X_{\text{trap}}$ and $X_{\text{object}}$ denote the positions of the optical trap and the micro-object, respectively, and $\mathbf{F}_{\text{OT}}$ denotes the corresponding optical force. The parameter $K$ represents the local trap stiffness in the near-trap region, $\delta$ defines the transition threshold, and $A$ and $C$ parameterize the far-field approximation. In the digital twin, the fitted optical force is applied to the Multi-Sphere Distributed Manipulation (MSDM) model-based trapping elements for force-aware motion simulation and haptic rendering ~\cite{zhang2020distributed}, which will be detailed in Section \ref{MODEL}. Quantitative and qualitative comparisons with reference experimental observations are presented in Section~\ref{sec:digitaltwin_accuracy} to assess whether the simulation reproduces representative microrobot motion behaviors under the tested settings.


\subsection{Pose and Depth Estimation of Microrobots}

This module estimates the pose and depth of the microrobot from OT images to support deployment-oriented digital-twin alignment and visual reconstruction. Given a 2D microscopic image, a trained deep learning model predicts a discrete pose and a continuous depth value (Fig.~\ref{fig-CVestimation}(a)). The network comprises a shared feature-extraction backbone and two task-specific heads: a classification head for pose estimation and a regression head for depth estimation. The estimated state is then combined with the commanded optical-trap coordinates and transmitted to Isaac Sim via ROS, enabling alignment of the digital twin with reference OT observations.

In Isaac Sim, the imported 3D microrobot model is updated using the estimated pose/depth state and trap configuration, enabling reconstruction of the microrobot-trap spatial relationship. In the present study, this module is evaluated independently through pose/depth estimation accuracy and representative reconstruction quality in Section~\ref{sec:reconstruction_accuracy}.

\subsection{Model-Based Haptic Feedback} \label{MODEL}

As shown in Fig.~\ref{fig-controlBlock}, the user manipulates the microrobot using two Geomagic Touch devices. Hand motions are converted into optical-trap motion commands in the digital twin through an incremental control scheme, and low-pass filtering is applied to suppress tremor and high-frequency jitter. Haptic feedback is generated from the model-based optical force estimated in the digital twin. Under the low-inertia microscale conditions considered here, the optical trapping force can be treated as a proxy for the resultant external load acting on the microrobot, because changes in viscous drag, contact interactions, or other disturbances alter the force required to maintain the relative position between the optical trap and the microrobot. The rendered force therefore provides the operator with an interpretable cue of interaction intensity and trapping stability during manipulation.


Following Zhang et al.~\cite{zhang2020distributed}, the MSDM model is adopted to approximate the complex-shaped microrobot as an assembly of spherical trapping elements. Within the digital twin, the reconstructed microrobot pose and trap configuration determine the relative positions between each trapping element and the optical traps. Based on the fitted 3D optical-force model, the haptic force \(\mathbf{F}_{\text{haptic}}\) is calculated as the vector sum of the optical forces acting on all trapping elements:

\begin{equation}
\mathbf{F}_{\text{haptic}} = \sum_{i=1}^{n} \mathbf{F}_{\text{OT}}(X_{\text{trap}}, X_{\text{ball}_i})
\label{eq:haptic_force}
\end{equation}


Here, $\mathbf{X}_{\mathrm{trap}}$ denotes the optical-trap position, and $\mathbf{X}_{\mathrm{ball}_i}$ denotes the position of the $i$-th spherical trapping element in the MSDM representation. The optical-force term $\mathbf{F}_{\mathrm{OT}}$ is computed using  Eq.~\eqref{eq:optical_force}. The forces acting on all trapping elements are summed to obtain the raw model-based force, which is then empirically scaled to match the output range of the Geomagic Touch haptic devices. To improve rendering stability and suppress high-frequency fluctuations caused by Brownian perturbations, abrupt contact changes, and hand tremor, low-pass filtering and virtual damping are applied in the haptic loop.


\begin{figure*}[htbp!]
  \captionsetup{font=footnotesize,labelsep=period}
\centering
\includegraphics[width=0.95\hsize]{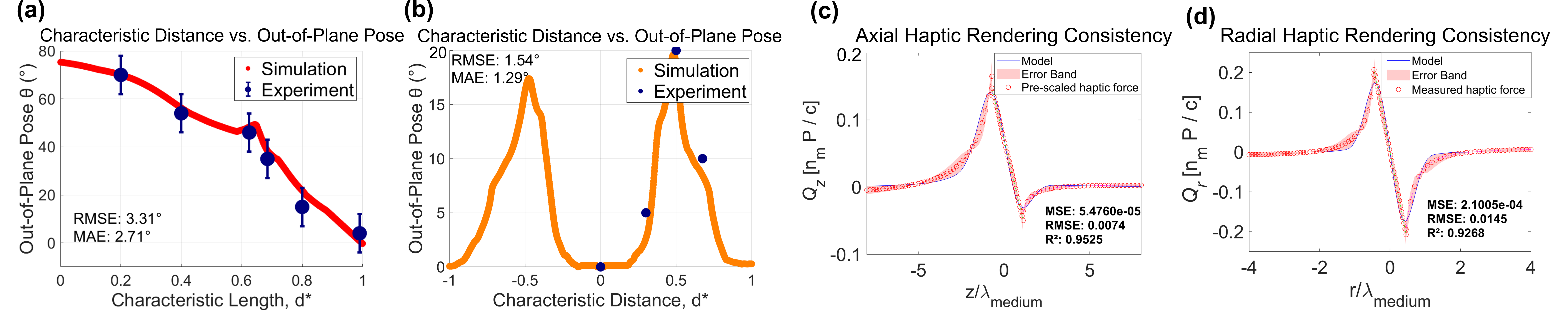}
\caption[Motion-behavior agreement and internal consistency of force rendering]{Validation of motion behavior and force-rendering consistency. \textbf{(a)} Out-of-plane rotation angle versus characteristic length \(d^*\) for Control Strategy A, comparing reference experiments and digital twin results. \textbf{(b)} Out-of-plane rotation angle versus characteristic distance \(d^*\) for Control Strategy B, comparing reference experiments and digital twin results. \textbf{(c)} Consistency between the fitted optical-force model and the pre-scaled haptic force along the axial direction. \textbf{(d)} Consistency between the fitted optical-force model and the pre-scaled haptic force along the radial direction.}
\label{fig-DigitalValidation}
    \vspace{-0.3cm}
\end{figure*}%

\section{Experiments and Results}
\label{experiments} 
\subsection{Evaluation of Motion Control Fidelity in Digital Twin} \label{sec:digitaltwin_accuracy}

\subsubsection{Experiment Description}

To assess whether the digital twin can reproduce representative motion behaviors observed in OT microrobot experiments, we adopted the microrobot configuration and reference control settings reported by Zhang et al.~\cite{zhang2020distributed}. In this comparison, two optical traps were applied simultaneously. Under Control Strategy A, the characteristic trap spacing \(d^*\) was varied to examine the resulting out-of-plane rotation of the microrobot. Under Control Strategy B, a power-distribution-based strategy with the trap-power distribution parameter fixed at \(m = 1.5\) was used, and the out-of-plane rotation was evaluated as a function of the characteristic distance \(d^*\). In both cases, the out-of-plane rotation angle \(\theta\) predicted by the digital twin was compared with the reference experimental results.

\subsubsection{Results and Analysis}

The results in Fig. \ref{fig-DigitalValidation}(a) and (b) show that the digital twin captures the main rotational behaviors reported in the reference OT experiments under both control settings. In Control Strategy A (Fig. \ref{fig-DigitalValidation}(a)), the digital twin follows the experimentally observed decrease in out-of-plane rotation angle with increasing \(d^*\), yielding an RMSE of 3.31$^{\circ}$ and a MAE of 2.71$^{\circ}$. In Control Strategy B (Fig. \ref{fig-DigitalValidation}(b)), the digital twin reproduces the characteristic rotation profile under the power-distribution-based setting, with an RMSE of 1.54$^{\circ}$ and a MAE of 1.29$^{\circ}$. These results indicate trend-level agreement with the reference data and support the use of the digital twin for motion analysis and controller development of OT-actuated microrobots.

\subsection{Internal Consistency of Model-Based Haptic Rendering}\label{sec:haptic_realism}

\subsubsection{Experiment Description}

To evaluate the consistency of the haptic rendering module, we compared the pre-scaled haptic force with the fitted optical-force model used in the digital twin. The comparison was carried out along representative axial and radial directions by sampling the pre-scaled haptic force and matching it to the corresponding fitted force curves. Fig. \ref{fig-DigitalValidation}(c) and (d) show the fitted optical-force model together with the pre-scaled rendered force in the two directions.

\subsubsection{Results and Analysis}

According to Fig.~\ref{fig-DigitalValidation}(c), the axial comparison yielded MSE = $5.460 \times 10^{-5}$, RMSE = 0.0074, and $R^2$ = 0.9525. For the radial direction in Fig.~\ref{fig-DigitalValidation}(d), the corresponding values were MSE = $2.1005 \times 10^{-4}$, RMSE = 0.0145, and $R^2$ = 0.9268. These results show that the pre-scaled rendered force closely follows the fitted optical-force model in the tested directions, supporting the numerical consistency of the force-rendering pipeline before empirical scaling for device-side haptic display. This agreement indicates that the rendered force channel can serve as a faithful proxy for conveying model-estimated interaction-force variations during manipulation. Together with the motion-agreement results in Section \ref{sec:digitaltwin_accuracy}, these results support the use of the digital twin for simulation-based analysis and development of visuo-haptic OT manipulation strategies.

\subsection{Evaluation of Microrobot Pose and Depth Estimation} \label{sec:reconstruction_accuracy}

\subsubsection{Experiment Description}

To evaluate image-based pose classification and depth estimation for digital-twin reconstruction, a microrobot fixed on a glass plane was driven by a piezoelectric stage to generate labeled pose and depth variations. Motion along the $z$-axis provided depth labels, while actuation about the $x$-axis generated 19 discrete out-of-plane poses characterized by $\theta$ (Fig.~\ref{fig-CVestimation}). A total of 5,700 microscope images were collected and split into 3,800 for training, 950 for validation, and 950 for testing. Pre-trained VGG16~\cite{simonyan2014very}, ResNet18~\cite{he2016deep}, and ResNet50~\cite{he2016deep} were used as backbones, each equipped with a pose-classification head and a depth-regression head. Each head comprised a linear layer reducing the feature dimension to 256, followed by ReLU and dropout. The classification head predicted the pose class via softmax, while the regression head estimated normalized depth. All models were fine-tuned for 20 epochs.

\subsubsection{Results and Analysis}

\begin{figure*}[t!]
  \captionsetup{font=footnotesize,labelsep=period}
\centering

\includegraphics[width=0.95\hsize]{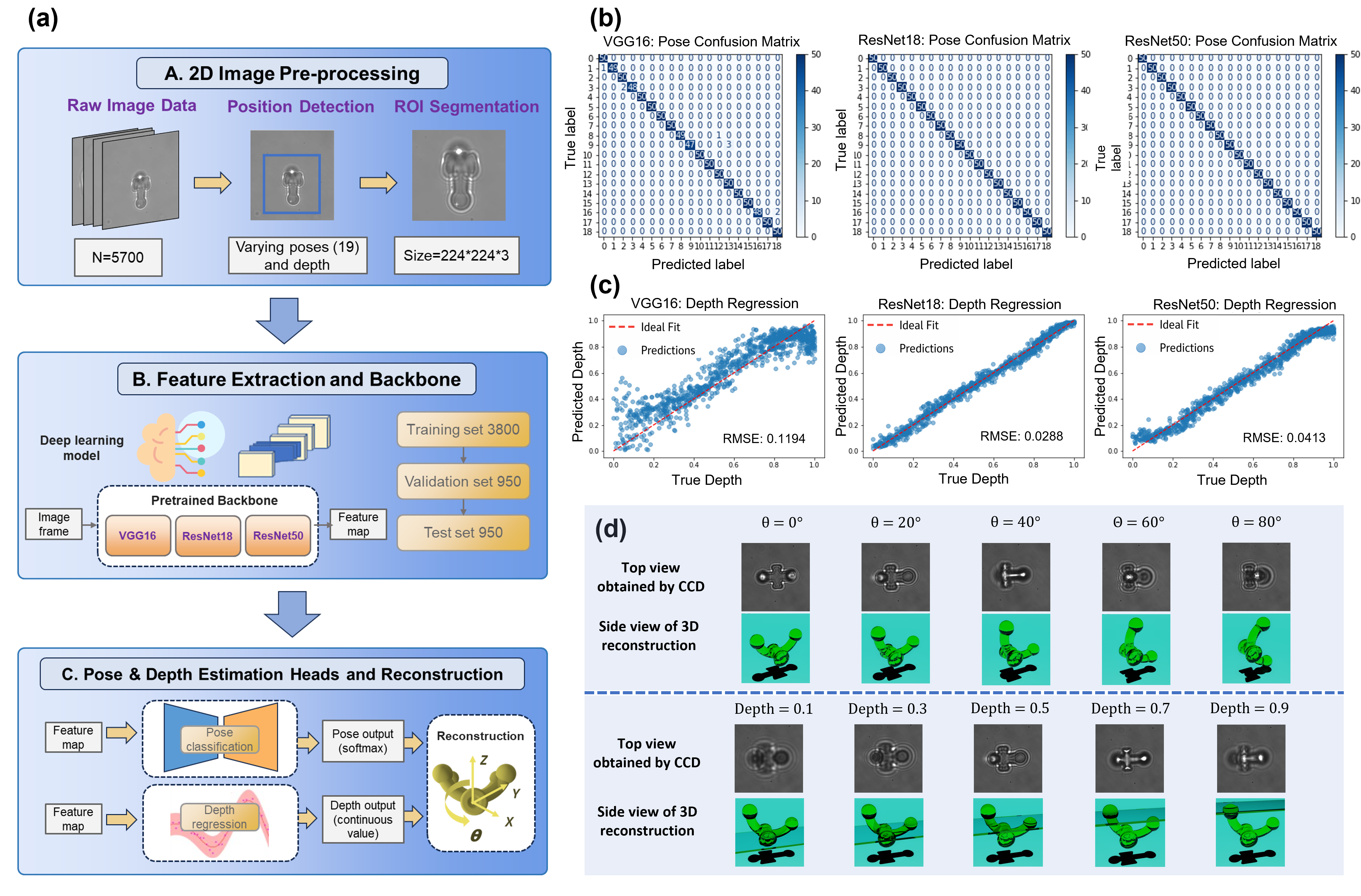}
 \vspace{-0.3cm}
\caption[Dataset generation, model comparison, and representative reconstruction results for microrobot pose classification and depth estimation]{Dataset generation, model comparison, and representative reconstruction results for microrobot pose classification and depth estimation. (a) Data-generation and network workflow for pose/depth estimation and digital-twin reconstruction. (b) Confusion matrices for pose classification using VGG16, ResNet18, and ResNet50. (c) Scatter plots of predicted versus ground-truth normalized depth for the three models. (d) Representative top-view microscope images and corresponding digital-twin reconstructions for different pose angles and normalized depths. For the pose-varying set, the normalized depth was fixed at 0.5. For the depth-varying set, the pose angle was fixed at $\theta = 20^\circ$.}
\label{fig-CVestimation}
    \vspace{-0.6cm}
\end{figure*}

Pose estimation was evaluated by accuracy and F1-score, and depth estimation by RMSE on normalized depth labels. As summarized in Table~\ref{table-class_reg} and Fig.~\ref{fig-CVestimation}(b)--(c), ResNet18 achieved the best overall performance, reaching 100\% accuracy and F1-score for pose classification and the lowest depth RMSE of 0.0288. ResNet50 matched the classification performance but yielded a higher depth RMSE of 0.0413, whereas VGG16 showed the largest depth dispersion and the highest RMSE of 0.1194. Representative results in Fig.~\ref{fig-CVestimation}(d) further show that the reconstructed side views vary consistently with the corresponding top-view microscope images across both pose-varying and depth-varying cases, indicating that the estimation module provides informative spatial cues for subsequent visualization and visuo-haptic interaction in the digital twin.

\begin{table}[!t]
\centering
\captionsetup{font=footnotesize,labelsep=period}
\caption{Comparison of different backbone models for pose classification and normalized-depth estimation. The best results are highlighted in \textbf{bold}.}
\begin{tabular}{c|ccc}
\hline
\multirow{2}{*}{Model} & \multicolumn{2}{c}{Pose}  & Depth \\
&Acc.($\uparrow$) & F1($\uparrow$) & RMSE($\downarrow$)\\
\hline
VGG16    &   0.99  &  0.99   &  0.1194 \\
ResNet18 &  \textbf{1.00}   &  \textbf{1.00}   &  \textbf{0.0288} \\ 
ResNet50 &  \textbf{1.00}   &  \textbf{1.00} &  0.0413 \\  
\hline
\end{tabular}
\label{table-class_reg}
\vspace{-0.5cm}
\end{table}

\subsection{User Evaluation of the Digital-Twin-Driven Haptic Teleoperation Interface}

\subsubsection{Experiment Description}

Five volunteers, aged 22-27, participated in the study. Each volunteer completed two simulated cell-delivery trials under each interface condition, resulting in a total of 10 trials per condition. In each trial, the operator guided the microrobot from a predefined start point to a predefined endpoint while avoiding trap loss and excessive interaction force. A trial was considered successful if the simulated cell reached the endpoint in a single attempt without losing the optical trap. Two interface conditions were evaluated: a haptic-feedback condition, which combined the visual interface with model-based force rendering, and a no-haptic-feedback condition, which used the same visual interface with force rendering disabled. During each trial, the system recorded the simulated cell contact force and the distance between the microrobot and the trap center (i.e., the microrobot-to-trap-center distance).

\subsubsection{Results and Analysis}

\begin{figure}[t!]
  \captionsetup{font=footnotesize,labelsep=period}
\centering
\includegraphics[width=1\columnwidth]{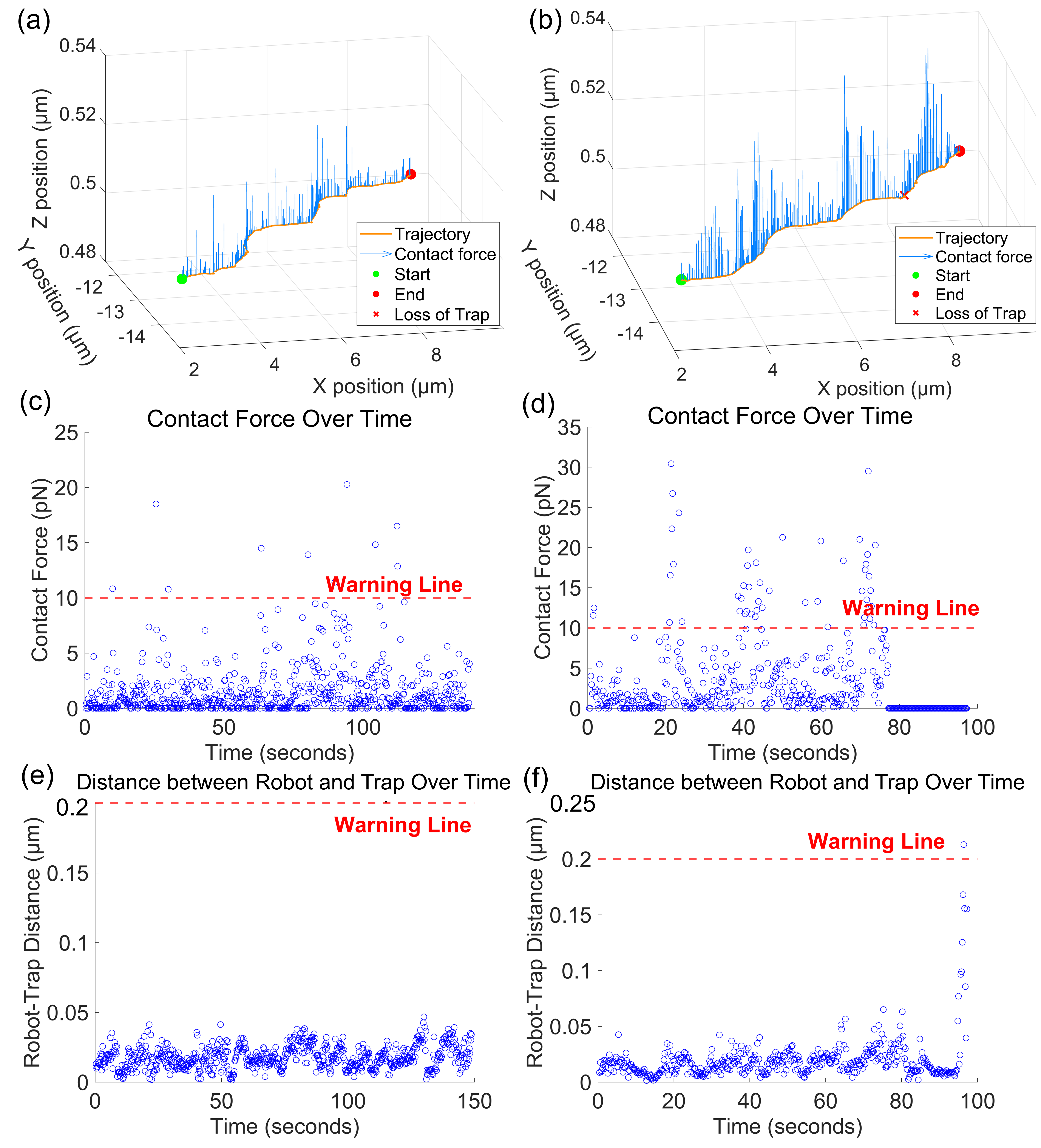}
\caption[3D trajectory with force vector visualization]{Representative task trajectories and temporal metrics from the user evaluation under two interface conditions: Panels (a), (c), and (e) correspond to the haptic-feedback condition, whereas panels (b), (d), and (f) correspond to the non-haptic condition. \textbf{(a)} and \textbf{(b)} Simulated cell-contact-force visualization along the transport path in the two conditions. \textbf{(c)} and \textbf{(d)} Temporal variation of simulated cell contact force in the two conditions. \textbf{(e)} and \textbf{(f)} Temporal variation of the distance between the microrobot and the optical-trap center in the two conditions.}
\label{fig-hapticValidation}
\vspace{-0.7cm}
\end{figure}

Aggregate results are summarized in Table~\ref{table-contact_force_distance}. With haptic feedback, the mean contact-force metric decreased from 5.39 to 2.57 and the mean distance metric from 0.04 to 0.02. Their standard deviations also decreased from 6.78 to 3.18 and from 0.058 to 0.026, corresponding to reductions of 53.2\% and 55.2\%, respectively. Task success improved from 30\% to 80\%.
Fig.~\ref{fig-hapticValidation}(a) and (b) show representative task trajectories under the two interface conditions. With model-based force rendering enabled, the transport path remained smoother and the simulated contact-force vectors were generally smaller, whereas the non-haptic condition exhibited larger force spikes and a trap-loss event near the end of the task. Consistently, Fig.~\ref{fig-hapticValidation}(c)-(f) show that haptic feedback produced fewer force excursions toward or beyond the preset warning threshold and kept the microrobot-to-trap-center distance more tightly clustered, while the non-haptic condition exhibited more dispersed force values and larger trap-loss-related deviations. Together, these results suggest that model-based haptic rendering can improve force regulation, interaction stability, and trap alignment in the evaluated simulated task.

\begin{table}[!h]
\centering
\captionsetup{font=footnotesize,labelsep=period} 
\caption{Comparison of Contact Force, robot-optical trap distance metric (Distance), and Task Success Rate with and without Haptic Feedback}
\begin{tabular}{|c|c|c|c|c|c|}
\hline
\multirow{2}{*}{\textbf{Mode}} & \multicolumn{2}{c|}{\textbf{Contact Force}} & \multicolumn{2}{c|}{\textbf{Distance}} & \multirow{2}{*}{\textbf{Success Rate}} \\
\cline{2-5}
 & Mean & SD & Mean & SD & \\
\hline
\textbf{With Haptic} & 2.57 & 3.18 & 0.02 & 0.026 & 80\% \\
\hline
\textbf{Without Haptic} & 5.39 & 6.78 & 0.04 & 0.058 & 30\% \\
\hline
\end{tabular}
\label{table-contact_force_distance}
\parbox{0.9\linewidth}{\scriptsize \textit{Note:} `SD' indicates standard deviation. `Success Rate' denotes the proportion of successful trials in the digital-twin-based task, where success means that the microrobot stably delivered the simulated cell to the predefined end point in a single attempt without trap loss. Each condition included 10 trials in total.}
\vspace{-0.4cm}
\end{table}

\section{Conclusions}
\label{conclusion}

This paper presented a digital twin framework for virtual visuo-haptic teleoperation of complex-shaped optical microrobots. The framework integrates bimanual interaction, optical-force-based microrobot motion simulation, 3D visual reconstruction, image-based pose/depth estimation, and model-based haptic rendering within a unified ROS-connected simulation environment. The digital twin reproduced representative microrobot motion trends, and the haptic rendering pipeline showed numerical consistency with the fitted optical-force model. In preliminary simulated cell-delivery tasks, haptic feedback reduced the standard deviations of the contact-force and microrobot-to-trap-center distance metrics by 53.2\% and 55.2\%, respectively, and improved task success from 30\% to 80\%.
These results demonstrate the potential of the proposed framework as a platform for developing and evaluating visuo-haptic teleoperation strategies for complex-shaped optical microrobots. 
Future work will extend the framework to more diverse manipulation tasks and physical OT deployment scenarios, with particular focus on real-time closed-loop integration and on evaluating how perception error and latency affect the accuracy and stability of haptic feedback.

\bibliographystyle{IEEEtran}
\bibliography{references}

@article{zhang2020distributed,
  title={Distributed force control for microrobot manipulation via planar multi-spot optical tweezer},
  author={Zhang, Dandan and Barbot, Antoine and Lo, Benny and Yang, Guang-Zhong},
  journal={Advanced Optical Materials},
  volume={8},
  number={21},
  pages={2000543},
  year={2020},
  publisher={Wiley Online Library}
}

@inproceedings{he2016deep,
  title={Deep residual learning for image recognition},
  author={He, Kaiming and Zhang, Xiangyu and Ren, Shaoqing and Sun, Jian},
  booktitle={Proceedings of the IEEE conference on computer vision and pattern recognition},
  pages={770--778},
  year={2016}
}

@article{simonyan2014very,
  title={Very deep convolutional networks for large-scale image recognition},
  author={Simonyan, Karen},
  journal={arXiv preprint arXiv:1409.1556},
  year={2014}
}

@article{guo2024lightweight,
  title={A lightweight and affordable wearable haptic controller for robot-assisted microsurgery},
  author={Guo, Xiaoqing and McFall, Finn and Jiang, Peiyang and Liu, Jindong and Lepora, Nathan and Zhang, Dandan},
  journal={Sensors},
  volume={24},
  number={9},
  pages={2676},
  year={2024},
  publisher={MDPI}
}

@article{tanaka2022dual, title={Dual-Arm Visuo-Haptic Optical Tweezers for Bimanual Cooperative Micromanipulation of Nonspherical Objects}, author={Tanaka, Yoshio and Fujimoto, Ken’ichi}, journal={Micromachines}, volume={13}, number={11}, pages={1830}, year={2022}, publisher={MDPI} }

@article{keshmiri2022digital, title={Digital twin of a magnetic medical microrobot with stochastic model predictive controller boosted by machine learning in cyber-physical healthcare systems}, author={Keshmiri Neghab, Hamid and Jamshidi, Mohammad and Keshmiri Neghab, Hamed}, journal={Information}, volume={13}, number={7}, pages={321}, year={2022}, publisher={MDPI} }

@preamble{"\newcommand{\CDC}[1]{Proceedings of the #1 IEEE Conference on Decision and Control}"}

@article{zhang2022fabrication,
  title={Fabrication and optical manipulation of micro-robots for biomedical applications},
  author={Zhang, Dandan and Ren, Yunxiao and Barbot, Antoine and Seichepine, Florent and Lo, Benny and Ma, Zhuo-Chen and Yang, Guang-Zhong},
  journal={Matter},
  volume={5},
  number={10},
  pages={3135--3160},
  year={2022},
  publisher={Elsevier}
}

@article{lee2021real,
  title={Real-time teleoperation of magnetic force-driven microrobots with 3D haptic force feedback for micro-navigation and micro-transportation},
  author={Lee, Jaeyeon and Zhang, Xiao and Park, Chung Hyuk and Kim, Min Jun},
  journal={IEEE Robotics and Automation Letters},
  volume={6},
  number={2},
  pages={1769--1776},
  year={2021},
  publisher={IEEE}
}

@article{nieminen2007optical,
  title={Optical tweezers computational toolbox},
  author={Nieminen, Timo A and Loke, Vincent LY and Stilgoe, Alexander B and Kn{\"o}ner, Gregor and Bra{\'n}czyk, Agata M and Heckenberg, Norman R and Rubinsztein-Dunlop, Halina},
  journal={Journal of Optics A: Pure and Applied Optics},
  volume={9},
  number={8},
  pages={S196},
  year={2007},
  publisher={IOP Publishing}
}

@inproceedings{raphalen2025haptic,
  title={Haptic shared control of a pair of microrobots for telemanipulation using constrained optimization},
  author={Raphalen, Leon and Ferro, Marco and Misra, Sarthak and Giordano, Paolo Robuffo and Pacchierotti, Claudio},
  booktitle={2025 IEEE/RSJ International Conference on Intelligent Robots and Systems (IROS)},
  pages={17391--17397},
  year={2025},
  organization={IEEE}
}

@incollection{grieves2016digital,
  title={Digital twin: Mitigating unpredictable, undesirable emergent behavior in complex systems},
  author={Grieves, Michael and Vickers, John},
  booktitle={Transdisciplinary perspectives on complex systems: New findings and approaches},
  pages={85--113},
  year={2016},
  publisher={Springer}
}

@article{volpe2023roadmap,
  title={Roadmap for optical tweezers},
  author={Volpe, Giovanni and Marag{\`o}, Onofrio M and Rubinsztein-Dunlop, Halina and Pesce, Giuseppe and Stilgoe, Alexander B and Volpe, Giorgio and Tkachenko, Georgiy and Truong, Viet Giang and Chormaic, S{\'\i}le Nic and Kalantarifard, Fatemeh and others},
  journal={Journal of Physics: Photonics},
  volume={5},
  number={2},
  pages={022501},
  year={2023},
  publisher={IOP Publishing}
}

@article{yadav2024optical,
  title={Optical tweezers in biomedical research--progress and techniques},
  author={Yadav, Dharm Singh and Savopol, Tudor},
  journal={Journal of Medicine and Life},
  volume={17},
  number={11},
  pages={978},
  year={2024}
}

@article{fan2023digital,
  title={Digital twin-driven mixed reality framework for immersive teleoperation with haptic rendering},
  author={Fan, Wen and Guo, Xiaoqing and Feng, Enyang and Lin, Jialin and Wang, Yuanyi and Liang, Jiaming and Garrad, Martin and Rossiter, Jonathan and Zhang, Zhengyou and Lepora, Nathan and others},
  journal={IEEE Robotics and Automation Letters},
  volume={8},
  number={12},
  pages={8494--8501},
  year={2023},
  publisher={IEEE}
}

@article{riaziat2024investigating,
  title={Investigating haptic feedback in vision-deficient millirobot telemanipulation},
  author={Riaziat, Naveed D and Erin, Onder and Krieger, Axel and Brown, Jeremy D},
  journal={IEEE robotics and automation letters},
  volume={9},
  number={7},
  pages={6178--6185},
  year={2024},
  publisher={IEEE}
}

@article{si2024enabling,
  title={Enabling intuitive and effective micromanipulation: A wearable exoskeleton-integrated macro-to-micro teleoperation system with a 3D electrothermal microgripper},
  author={Si, Guoning and Zhang, Hanli and Zhang, Zhuo and Zhang, Xuping},
  journal={Robotics and Autonomous Systems},
  volume={181},
  pages={104776},
  year={2024},
  publisher={Elsevier}
}

@article{hu2022advanced,
  title={Advanced optical tweezers on cell manipulation and analysis},
  author={Hu, Sheng and Ye, Jun-yan and Zhao, Yong and Zhu, Cheng-liang},
  journal={The European Physical Journal Plus},
  volume={137},
  number={9},
  pages={1024},
  year={2022},
  publisher={Springer}
}

@incollection{gerena20233d,
  title={3D force-feedback optical tweezers for experimental biology},
  author={Gerena, Edison and Haliyo, Sinan},
  booktitle={Robotics for cell manipulation and characterization},
  pages={145--172},
  year={2023},
  publisher={Elsevier}
}

@article{pan2026optical,
  title={Optical manipulation-based cell modulation and bio-microrobot},
  author={Pan, Ting and Bi, Aoquan and Xin, Hongbao},
  journal={Journal of Innovative Optical Health Sciences},
  year={2026},
  publisher={World Scientific}
}

@inproceedings{tan2025interactive,
  title={Interactive OT Gym: A Reinforcement Learning-Based Interactive Optical tweezer (OT)-Driven Microrobotics Simulation Platform},
  author={Tan, Zongcai and Zhang, Dandan},
  booktitle={2025 IEEE International Conference on Robotics and Automation (ICRA)},
  pages={1--7},
  year={2025},
  organization={IEEE}
}

@article{tan2025physics,
  title={Physics-Informed Machine Learning for Efficient Sim-to-Real Data Augmentation in Micro-Object Pose Estimation},
  author={Tan, Zongcai and Wei, Lan and Zhang, Dandan},
  journal={arXiv preprint arXiv:2511.16494},
  year={2025}
}
\end{document}